\documentclass{article}

\usepackage[numbers,sort&compress]{natbib}

\usepackage[dvipsnames]{xcolor}

\usepackage[final]{neurips_2019}

\usepackage[utf8]{inputenc} %
\usepackage[T1]{fontenc}    %
\usepackage{url}            %
\usepackage{booktabs}       %
\usepackage{amsfonts}       %
\usepackage{nicefrac}       %
\usepackage{microtype}      %
\usepackage{subcaption}
\usepackage{comment}
\usepackage{amsmath, amsfonts, bbm, amsthm, graphicx, enumitem}

\newtheorem{proposition}{Proposition}

\usepackage{algorithm, algorithmicx, algpseudocode}
\usepackage{enumitem}

\title{Bias Correction of Learned Generative Models using Likelihood-Free Importance Weighting}

\author{%
Aditya Grover$^1$, Jiaming Song$^1$, Alekh Agarwal$^2$, Kenneth Tran$^2$, \\
\textbf{Ashish Kapoor$^2$, Eric Horvitz$^2$, Stefano Ermon$^1$}
 \\
  $^1$Stanford University, $^2$Microsoft Research, Redmond
}

\usepackage{amsmath,amsfonts,bm}

\def\1{\bm{1}}
\newcommand{\train}{D_{\mathrm{train}}}

\newcommand{\test}{D_{\mathrm{test}}}

\def\fid{{\textnormal{FID}}}

\def\calX{{\mathcal{X}}}
\def\calY{{\mathcal{Y}}}
\def\calZ{{\mathcal{Z}}}

\def\va{{\mathbf{a}}}
\def\vs{{\mathbf{s}}}
\def\vx{{\mathbf{x}}}
\def\vy{{\mathbf{y}}}
\def\vz{{\mathbf{z}}}

\DeclareMathAlphabet{\mathsfit}{\encodingdefault}{\sfdefault}{m}{sl}
\SetMathAlphabet{\mathsfit}{bold}{\encodingdefault}{\sfdefault}{bx}{n}

\def\gA{{\mathcal{A}}}

\def\gS{{\mathcal{S}}}

\newcommand{\pdata}{p_{\rm{data}}}

\newcommand{\ptheta}{p_{\theta}}

\newcommand{\E}{\mathbb{E}}

\newcommand{\KL}{D_{\mathrm{KL}}}

\begin{document}

\maketitle
\begin{abstract}
A learned generative model often produces biased statistics relative to the underlying data distribution. A standard technique to correct this bias is importance sampling, where samples from the model are weighted  by the likelihood ratio under model and true distributions. When the likelihood ratio is unknown, it can be estimated by training a probabilistic classifier to distinguish samples from the two distributions. We employ this likelihood-free importance weighting method to correct for the bias in generative models. We find that this technique consistently improves standard goodness-of-fit metrics for evaluating the sample quality of state-of-the-art deep generative models, suggesting reduced bias. Finally, we demonstrate its utility on representative applications in a) data augmentation for classification using generative adversarial networks, and b) model-based policy evaluation using off-policy data.
\end{abstract}
\section{Introduction}

Learning generative models of complex environments from high-dimensional observations is a long-standing challenge in machine learning. Once learned, these models are used to draw inferences and to plan future actions. For example, in data augmentation, samples from a learned model are used to enrich a dataset for supervised learning~\citep{antoniou2017data}. In model-based off-policy policy evaluation (henceforth MBOPE), a learned dynamics model is used to simulate and evaluate a target policy without real-world deployment~\citep{mannor2007bias}, which is especially valuable for risk-sensitive applications~\citep{thomas2015safe}.
In spite of the recent successes of deep generative models, existing theoretical results show that learning distributions in an unbiased manner is either impossible or has prohibitive sample complexity~\citep{rosenblatt1956remarks, arora2018gans}. Consequently, the models used in practice are inherently \emph{biased},\footnote{We call a generative model biased if it produces biased statistics relative to the true data distribution.} and 
can lead to misleading downstream inferences.

In order to address this issue, we start from the observation that many typical uses of generative models involve computing expectations under the model. 
For instance, in MBOPE, we seek to find the expected return of a policy under a trajectory distribution defined by this policy and a learned dynamics model.
A classical recipe for correcting the bias in expectations, when samples from a different distribution than the ground truth are available, is to importance weight the samples according to the likelihood ratio~\citep{HorvitzTh52}.
If the importance weights were exact, the resulting estimates are unbiased. But in practice, the likelihood ratio is unknown and needs to be estimated since the true data distribution is unknown and even the model likelihood is intractable or ill-defined for many deep generative models, e.g., variational autoencoders~\citep{kingma2013auto} and generative adversarial networks~\citep{goodfellow2014generative}.%

Our proposed solution to estimate the importance weights is to train a calibrated, probabilistic classifier to distinguish samples from the data distribution and the generative model. 
As shown in prior work, the output of such classifiers can be used to extract density ratios~\citep{sugiyama2012density}.
Appealingly, this estimation procedure is likelihood-free since it only requires samples from the two distributions. 

Together, the generative model and the importance weighting function (specified via a binary classifier) induce a new unnormalized distribution.
While exact density estimation and sampling from this induced distribution is intractable, we can derive a particle based approximation which permits efficient sampling via resampling based methods.
We derive conditions on the quality of the weighting function such that the induced distribution provably improves the fit to the the data distribution. 

Empirically, we evaluate our bias reduction framework on three main sets of experiments. 
First, we consider goodness-of-fit metrics for evaluating sample quality metrics of a likelihood-based and a likelihood-free state-of-the-art (SOTA) model on the CIFAR-10 dataset.
All these metrics are defined as Monte Carlo estimates from the generated samples.
By importance weighting samples, we observe a bias reduction of 23.35\% and 13.48\% averaged across commonly used sample quality metrics on PixelCNN++~\citep{salimans2017pixelcnn++} and SNGAN~\citep{miyato2018spectral} models respectively.

Next, we demonstrate the utility of our approach on the task of data augmentation for multi-class classification on the Omniglot dataset~\citep{lake2015human}. 
We show that, while naively extending the model with samples from a data augmentation, a generative adversarial network \citep{antoniou2017data} is not very effective for multi-class classification, we can improve classification accuracy from 66.03\% to 68.18\% by importance weighting the contributions of each augmented data point.

Finally, we demonstrate bias reduction for MBOPE~\citep{precup2000eligibility}. 
A typical MBOPE approach is to first estimate a generative model of the dynamics using off-policy data and then evaluate the policy via Monte Carlo~\citep{mannor2007bias,thomas2016data}. 
Again, we observe that correcting the bias of the estimated dynamics model via importance weighting reduces RMSE for MBOPE by 50.25\% on 3 MuJoCo environments~\citep{todorov2012mujoco}.

\section{Preliminaries}\label{sec:prelim}

\noindent \textbf{Notation.} Unless explicitly stated otherwise, we assume that probability distributions admit absolutely continuous densities on a suitable reference measure.
We use uppercase notation $X, Y, Z$  to denote random variables and lowercase notation $x, y, z$ to denote specific values in the corresponding sample spaces $\calX, \calY, \calZ$. 
We use boldface for multivariate random variables and their vector values.

\noindent \textbf{Background.} Consider a finite dataset $\train$ of instances $\vx$ drawn i.i.d. from a fixed (unknown) distribution $\pdata$. 
Given $\train$, the goal of generative modeling is to learn a distribution $\ptheta$ to approximate $\pdata$. Here, $\theta$ denotes the model parameters, e.g. weights in a neural network for deep generative models.
The parameters can be learned via maximum likelihood estimation (MLE) as in the case of autoregressive models~\citep{uria2016neural}, normalizing flows~\citep{dinh2014nice}, and variational autoencoders~\citep{kingma2013auto,rezende2014stochastic}, or via adversarial training e.g., using generative adversarial networks~\citep{goodfellow2014generative,mohamed2016learning} and variants.

\begin{comment}
Broadly, there exist two main paradigms for learning a deep generative model: maximum likelihood estimation (MLE) and adversarial training~\citep{mohamed2016learning}. MLE fits parameters to maximize the model likelihood for the dataset $\train$, when the model specifies a likelihood function, e.g. autoregressive models~\citep{uria2016neural}, normalizing flow models~\citep{dinh2014nice}, and variational autoencoder models~\citep{kingma2013auto}. 
Adversarial training, on the other hand, learns a generative model via a minimax game between the generative model and an auxiliary critic, where the critic distinguishes the samples in $\train$ and from those generated by the model~\citep{goodfellow2014generative}. 
This method is likelihood-free since the learning objective only requires evaluating expectations w.r.t. the current model distribution during training, which can be done by drawing samples from the model.
\end{comment}

\noindent \textbf{Monte Carlo Evaluation}
We are interested in use cases where the goal is to evaluate or optimize expectations of functions under some distribution $p$ (either equal or close to the data distribution $\pdata$). 
Assuming access to samples from $p$ as well some generative model $p_\theta$, one extreme is to evaluate the sample average using the samples from $p$ alone. 
However, this ignores the availability of $p_\theta$, through which we have a virtually unlimited access of generated samples
ignoring computational constraints and hence, could improve the accuracy of our estimates when $p_\theta$ is close to $p$. 
We begin by presenting a direct motivating use case of data augmentation using generative models for training classifiers which generalize better.

\noindent \textbf{Example Use Case: } 
Sufficient labeled training data for learning classification and regression system is often expensive to obtain or susceptible to noise. \textit{Data augmentation} seeks to overcome this shortcoming by artificially injecting new datapoints into the training set. These new datapoints are derived from an existing labeled dataset, either by manual transformations (e.g., rotations, flips for images), or alternatively, learned via a generative model~\citep{ratner2017learning,antoniou2017data}.

Consider a supervised learning task over a labeled dataset $D_{\rm{cl}}$.
The dataset consists of feature and label pairs $(\vx, \vy)$, each of which is assumed to be sampled independently from a data distribution $\pdata(\vx, \vy)$ defined over $\calX \times \calY$. 
Further, let $\calY \subseteq \mathbb{R}^k$. 
In order to learn a classifier $f_\psi:\mathcal{X}\to \mathbb{R}^k$ with parameters $\psi$, we minimize the expectation of a loss $\ell: \calY \times \mathbb{R}^k\to \mathbb{R}$ over the dataset $D_{\rm{cl}}$:

\begin{align}\label{eq:classify}
\E_{\pdata(\vx, \vy)}[\ell(\vy, f_\psi(\vx))]\approx \frac{1}{\vert D_{\rm{cl}} \vert} \sum_{(\vx, \vy) \sim D_{\rm{cl}}} \mathcal{\ell}(\vy, f_\psi(\vx)).
\end{align}
E.g., $\ell$ could be 
the 
cross-entropy loss. 
A generative model for the task of data augmentation learns a joint distribution $\ptheta(\vx, \vy)$. 
Several algorithmic variants exist for learning the model's joint distribution and we defer the specifics %
to the experiments section.
Once the generative model is learned, it can be used to optimize the expected classification loss in Eq.~\eqref{eq:classify} under a mixture distribution of empirical data distributions and generative model distributions given as:
 \begin{align}\label{eq:data_aug}
 p_{\rm{mix}}(\vx, \vy) = m \pdata(\vx, \vy) + (1-m) p_\theta(\vx, \vy)
\end{align}
for a suitable choice of the mixture weights $m\in [0,1]$. 
Notice that, while the eventual task here is optimization, reliably evaluating the expected loss of a candidate parameter $\psi$ is an important ingredient. We focus on this basic question first in advance of leveraging the solution for data augmentation. 
Further, even if evaluating the expectation once is easy, optimization requires us to do repeated evaluation (for different values of $\psi$) which is significantly more challenging.
Also observe that the distribution $p$ under which we seek expectations is same as $\pdata$ here, and we rely on the generalization of $p_\theta$ to generate transformations of an instance in the dataset which are not explicitly present, but plausibly observed in other, similar instances~\citep{zhao2018bias}.

\section{Likelihood-Free Importance Weighting}\label{sec:lfiw}

Whenever the distribution $p$, under which we seek expectations, differs from $\ptheta$, model-based estimates exhibit bias.
In this section, we start out by formalizing bias for Monte Carlo expectations and subsequently propose a bias reduction strategy based on likelihood-free importance weighting (LFIW). 
We are interested in evaluating expectations of a class of functions of interest $f \in \mathcal{F}$ w.r.t. the distribution $p$. For any given $f: \mathcal{X} \to \mathbb{R}$, we have $\E_{\vx \sim p}[f(\vx)] = \int p(\vx) f(\vx) \mathrm{d} \vx$.

Given access to samples from a generative model $p_\theta$, if we knew the densities for both $p$ and $p_\theta$, then a classical scheme to evaluate expectations under $p$ using samples from $p_\theta$ is to use importance sampling~\citep{HorvitzTh52}. We reweight each sample from $p_\theta$ according to its likelihood ratio under $p$ and $p_\theta$ and compute a weighted average of the function $f$ over these samples. 
\begin{align}
    \E_{\vx \sim p}[f(\vx)] &= \E_{\vx \sim p_{\theta}}\left[\frac{p(\vx)}{p_{\theta}(\vx)}f(\vx)\right]  \approx \frac{1}{T}\sum_{i=1}^T w(\vx_i) f(\vx_i)\label{eq:is}
\end{align}
where $w(\vx_i):=\nicefrac{p(\vx_i)}{p_{\theta}(\vx_i)}$ is the importance weight for $\vx_i \sim p_\theta$. The validity of this procedure is subject to the use of a proposal $p_\theta$ such that for all $\vx \in \calX$ where $p_\theta(\vx)= 0$, we also have $f(\vx)p(\vx)=0$.\footnote{A stronger sufficient, but not necessary condition that is independent of $f$, states that the proposal $p_\theta$ is valid if it has a support larger than $p$, i.e., for all $\vx \in \calX$, $p_\theta(\vx)= 0$ implies $p(\vx) = 0$.}

To apply this technique to reduce the bias of a generative sampler $\ptheta$ w.r.t. $p$, we require knowledge of the importance weights $w(\vx)$ for any $\vx \sim \ptheta$. However, we typically only have a sampling access to $p$ via finite datasets. 
For instance, in the data augmentation example above, where $p = \pdata$, the unknown distribution used to learn $\ptheta$. 
Hence we need a scheme to learn the weights $w(\vx)$, using samples from $p$ and $\ptheta$, which is the problem we tackle next.%
In order to do this, we consider a binary classification problem over $\calX \times \calY$ where $\calY=\{0, 1\}$ and the joint distribution is denoted as $q(\vx,y)$.
Let $\gamma=\frac{q(y=0)}{q(y=1)}>0$ denote any fixed odds ratio.
To specify the joint $q(\vx,y)$, we additionally need the conditional $q(\vx\vert y)$ which we define as follows:
\begin{align}
    q(\vx\vert y) = 
    \begin{cases}
      p_\theta(\vx) \text{ if }y=0\\
      p(\vx) \text{ otherwise}.
    \end{cases}
\end{align}

Since we only assume sample access to $p$ and $\ptheta(\vx)$, our strategy would be to estimate the conditional above via \textit{learning} a probabilistic binary classifier.
To train the classifier, we only require datasets of samples from $\ptheta(\vx)$ and $p(\vx)$ and estimate $\gamma$ to be the ratio of the size of two datasets.
Let $c_\phi: \calX \rightarrow [0, 1]$ denote the probability assigned by the classifier with parameters $\phi$ to a sample $\vx$ belonging to the positive class $y=1$. 
As shown in prior work~\citep{sugiyama2012density,grover2018boosted}, if $c_\phi$ is Bayes optimal, then the importance weights can be obtained via this classifier as:
\begin{align}
 w_\phi (\vx) = \frac{p(\vx)}{\ptheta(\vx)}= \gamma \frac{c_\phi(\vx)}{1-c_\phi(\vx)}.
\end{align}

In practice, we do not have access to a Bayes optimal classifier and hence, the estimated importance weights will not be exact. Consequently, we can hope to reduce the bias as opposed to eliminating it entirely. 
Hence, our default LFIW estimator is given as:
\begin{align}\label{eq:iw_default}
 \E_{\vx \sim p}[f(\vx)] \approx  \frac{1}{T}\sum_{i=1}^T \hat{w}_\phi(\vx_i) f(\vx_i)
\end{align}
where $\hat{w}_\phi(\vx_i) = \gamma \frac{c_\phi(\vx_i)}{1-c_\phi(\vx_i)}$ is the importance weight for $\vx_i \sim \ptheta$ estimated via $c_\phi(\vx)$. 

\textbf{Practical Considerations.}
\label{sec:practical_lfiw}
Besides imperfections in the classifier, the quality of a generative model also dictates the efficacy of importance weighting. 
For example, images generated by deep generative models often possess distinct artifacts which can be exploited by the classifier to give highly-confident predictions~\citep{odena2016deconvolution,odena2019open}. 
This could lead to very small importance weights for some generated images, and consequently greater relative variance in the importance weights across the Monte Carlo batch.
Below, we present some practical variants of LFIW estimator to offset this challenge. 

\begin{enumerate}[leftmargin=0cm,itemindent=.5cm,labelwidth=\itemindent,labelsep=0cm,align=left]
    \item \textit{Self-normalization:} The self-normalized LFIW estimator for Monte Carlo evaluation normalizes the importance weights across a sampled batch:
    \begin{align}\label{eq:self_norm_iw}
    \E_{\vx \sim p}[f(\vx)] \approx \sum_{i=1}^T \frac{\hat{w}_\phi(\vx_i)}{\sum_{j=1}^T \hat{w}_\phi(\vx_j)} f(\vx_i) \text{  where } \vx_i \sim \ptheta.
    \end{align}
    \item \textit{Flattening:} The flattened LFIW estimator interpolates between the uniform importance weights and the default LFIW weights via a power scaling parameter $\alpha \geq 0$:
    \begin{align}\label{eq:flattened_lfiw}
        \E_{\vx \sim p}[f(\vx)] \approx  \frac{1}{T}\sum_{i=1}^T \hat{w}_\phi(\vx_i)^\alpha f(\vx_i) \text{  where } \vx_i \sim \ptheta.
    \end{align}
   For $\alpha=0$, there is no bias correction, and $\alpha=1$ returns the default estimator in Eq.~\eqref{eq:iw_default}. 
   For intermediate values of $\alpha$, we can trade-off bias reduction with any undesirable variance introduced.
    \item\textit{Clipping:} The clipped LFIW estimator specifies a lower bound $\beta \geq 0$ on the importance weights:
    \begin{align}\label{eq:clipped_lfiw}
        \E_{\vx \sim p}[f(\vx)] \approx  \frac{1}{T}\sum_{i=1}^T \max(\hat{w}_\phi(\vx_i), \beta) f(\vx_i) \text{  where } \vx_i \sim \ptheta.
    \end{align}
    When $\beta=0$, we recover the default LFIW estimator in Eq.~\eqref{eq:iw_default}. Finally, we note that these estimators are not exclusive and can be combined e.g., flattened or clipped weights can be normalized.
\end{enumerate}

\textbf{Confidence intervals.}
Since we have real and generated data coming from a finite dataset and parametric model respectively, we propose a combination of empirical and parametric bootstraps to derive confidence intervals around the estimated importance weights. See Appendix~\ref{app:confidence_intervals} for details.

\begin{figure*}[t]
    \centering
    \begin{subfigure}[b]{0.24\textwidth}
    \centering
    \includegraphics[width=\columnwidth]{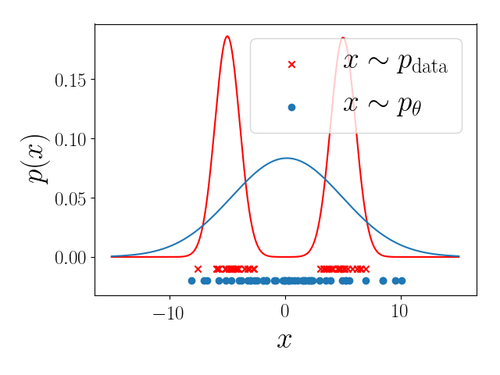}
    \caption{Setup}
    \end{subfigure}
    \begin{subfigure}[b]{0.24\textwidth}
    \centering
    \includegraphics[width=\columnwidth]{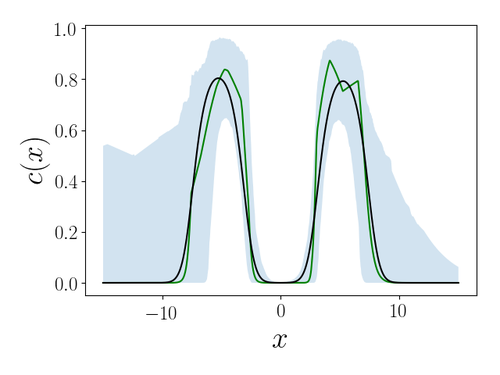}
    \caption{$n=50$}
    \end{subfigure}
    \begin{subfigure}[b]{0.24\textwidth}
    \centering
    \includegraphics[width=\columnwidth]{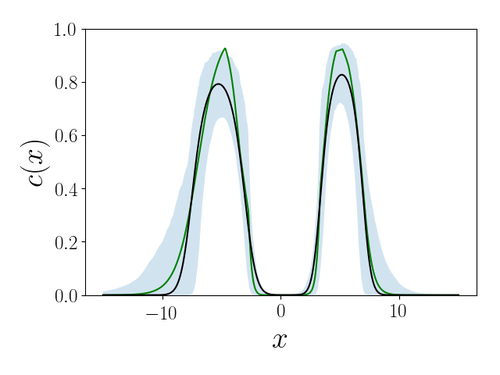}
    \caption{$n=100$}
    \end{subfigure}
    \begin{subfigure}[b]{0.24\textwidth}
    \centering
    \includegraphics[width=\columnwidth]{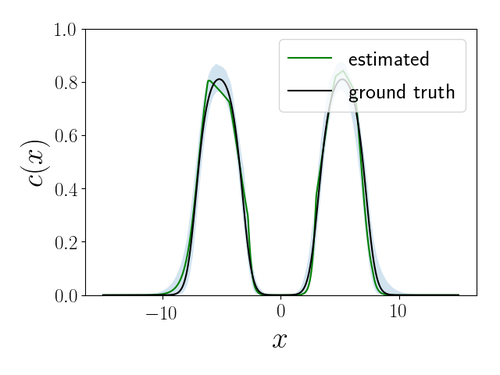}
    \caption{$n=1000$}
    \end{subfigure}
    \caption{Importance Weight Estimation using Probabilistic Classifiers. (a) A univariate Gaussian (blue) is fit to samples from a mixture of two Gaussians (red). (b-d) Estimated class probabilities (with 95\% confidence intervals based on $1000$ bootstraps) for varying number of points $n$, where $n$ is the number of points used for training the generative model and multilayer perceptron.}
    \label{fig:toy}
    \vspace{-0.1in}
\end{figure*} 

\textbf{Synthetic experiment.} We visually illustrate our importance weighting approach in a toy experiment
(Figure~\ref{fig:toy}a).
We are given a finite set of samples drawn from a mixture of two Gaussians (red). 
The model family is a unimodal Gaussian, illustrating mismatch due to a parametric model. 
The mean and variance of the model are estimated by the empirical means and variances of the observed data.
Using estimated model parameters, we then draw samples from the model (blue).

In Figure~\ref{fig:toy}b, we show the probability assigned by a binary classifier to a point to be from true data distribution. 
Here, the classifier is a single hidden-layer multi-layer perceptron.
The classifier is not Bayes optimal, which can be seen by the gaps between the optimal probabilities curve (black) and the estimated class probability curve (green).
However, as we increase the number of real and generated examples $n$ in Figures~\ref{fig:toy}c-d,
the classifier approaches optimality.
Furthermore, even its uncertainty shrinks with increasing data, as expected.
In summary, this experiment demonstrates 
how a binary classifier can mitigate this bias due to a mismatched generative model.

\begin{algorithm}[t]
   \caption{SIR for the Importance Resampled Generative Model $p_{\theta,\phi}$}
   \label{alg:sir}
\hspace*{\algorithmicindent} \textbf{Input:} Generative Model $\ptheta$, Importance Weight Estimator $\hat{w}_\phi$, budget $T$
\begin{algorithmic}[1]
\State Sample $\vx_1, \vx_2, \ldots, \vx_T$ independently from $\ptheta$
\State Estimate importance weights $\hat{w}(\vx_1), \hat{w}(\vx_2), \ldots, \hat{w}(\vx_T)$
\State Compute $\hat{Z}\leftarrow \sum_{t=1}^T \hat{w}(\vx_t)$
\State Sample $j \sim \textrm{Categorical}\left(\frac{\hat{w}(\vx_1)}{\hat{Z}}, \frac{\hat{w}(\vx_2)}{\hat{Z}}, \ldots, \frac{\hat{w}(\vx_T)}{\hat{Z}}\right)$
\State \Return $\vx_j$
\end{algorithmic}
\end{algorithm}
\section{Importance Resampled Generative Modeling}

In the previous section, we described a procedure to augment any base generative model $\ptheta$ with an importance weighting estimator $\hat{w}_\phi$ for debiased Monte Carlo evaluation.
Here, we will use this augmentation to induce an \textit{importance resampled generative model} with density $p_{\theta,\phi}$ given as:
\begin{align}\label{eq:induced_dist}
    p_{\theta,\phi}(\vx) \propto \ptheta(\vx) \hat{w}_\phi(\vx)
\end{align}
where the partition function is expressed as $Z_{\theta, \phi}=\int \ptheta(\vx) \hat{w}_\phi(\vx) \mathrm{d}\vx = \E_{\ptheta}[\hat{w}_\phi(\vx)]$.

\textbf{Density Estimation.} Exact density estimation requires a handle on the density of the base model $\ptheta$ (typically intractable for models such as VAEs and GANs) and estimates of the partition function.
Exactly computing the partition function is intractable.
If $\ptheta$ permits fast sampling and importance weights are estimated via LFIW (requiring only a forward pass through the classifier network), we can obtain unbiased estimates via a Monte Carlo average, i.e., $Z_{\theta, \phi} \approx \frac{1}{T} \sum_{i=1}^T \hat{w}_\phi(\vx_i)$ where $\vx_i\sim \ptheta$.
To reduce the variance, a potentially large number of samples are required. Since samples are obtained independently, the terms in the Monte Carlo average can be evaluated in parallel.

\textbf{Sampling-Importance-Resampling.}
While exact sampling from $p_{\theta,\phi}$ is intractable, we can instead perform sample from a particle-based approximation to $p_{\theta,\phi}$ via sampling-importance-resampling~\citep{liu1998sequential,doucet2000sequential} (SIR). 
We define the SIR approximation to $p_{\theta,\phi}$ via the following density:
\begin{align}\label{eq:sir_approx}
    p_{\theta,\phi}^{\mathrm{SIR}}(\vx; T) &:= \E_{\vx_2, \vx_3, \ldots, \vx_T \sim \ptheta} \left[ \frac{\hat{w}_\phi(\vx)}{\hat{w}_\phi(\vx) + \sum_{i=2}^T \hat{w}_\phi(\vx_i)}\ptheta(\vx)\right]
\end{align}
where $T>0$ denotes the number of independent samples (or ``particles").
For any finite $T$, sampling from $p_{\theta,\phi}^{\mathrm{SIR}}$ is tractable, as summarized
in Algorithm~\ref{alg:sir}.
Moreover, any expectation w.r.t. the SIR approximation to the induced distribution can be evaluated in closed-form using the self-normalized LFIW estimator (Eq.~\ref{eq:self_norm_iw}).
In the limit of $T\to\infty$, we recover the induced distribution $p_{\theta,\phi}$:
\begin{align}
\lim_{T\to\infty} p_{\theta,\phi}^{\mathrm{SIR}}(\vx; T) = p_{\theta,\phi}(\vx) \;\;\; \forall \vx
\end{align}

Next, we analyze conditions under which the resampled density $p_{\theta,\phi}$ provably improves the model fit to $\pdata$. 
In order to do so, we further assume that $\pdata$ is absolutely continuous w.r.t. $\ptheta$ and $p_{\theta,\phi}$.
We define the change in KL via the importance resampled density as:
\begin{align}\label{eq:delta_kl}
    \Delta(\pdata, \ptheta, p_{\theta,\phi}) &:= \KL(\pdata, p_{\theta,\phi})  - \KL(\pdata, \ptheta).
\end{align}

Substituting Eq.~\ref{eq:induced_dist} in Eq.~\ref{eq:delta_kl}, we can simplify the above quantity as:
\begin{align}
    \Delta(\pdata, \ptheta, p_{\theta,\phi}) &= \E_{\vx \sim \pdata}[- \log (\ptheta(\vx) \hat{w}_\phi(\vx))  + \log Z_{\theta, \phi} + \log \ptheta(\vx) ] \\
    &=\E_{\vx \sim \pdata}[\log \hat{w}_\phi(\vx)] - 
    \log \E_{\vx \sim \ptheta}[ \hat{w}_\phi(\vx)].\label{eq:nec_suf}
\end{align}

The above expression provides a necessary and sufficient condition for any positive real valued function (such as the LFIW classifier in Section~\ref{sec:lfiw}) to improve the KL divergence fit to the underlying data distribution. 
In practice, an unbiased estimate of the LHS can be obtained via Monte Carlo averaging of log- importance weights based on $\train$. 
The empirical estimate for the RHS is however biased.\footnote{If $\hat{Z}$ is an unbiased estimator for $Z$, then $\log\hat{Z}$ is a biased estimator for $\log Z$ via Jensen's inequality.}
To remedy this shortcoming, we consider the following necessary but insufficient condition.

\begin{proposition}\label{thm:nec}
If $\Delta(\pdata, \ptheta, p_{\theta,\phi})\geq 0$, then the following conditions hold:
\begin{align}
\E_{\vx \sim \pdata}[\hat{w}_\phi(\vx)] &\geq \E_{\vx \sim \ptheta}[ \hat{w}_\phi(\vx)]\label{eq:nec1},\\
\E_{\vx \sim \pdata}[\log \hat{w}_\phi(\vx)] &\geq \E_{\vx \sim \ptheta}[ \log \hat{w}_\phi(\vx)]\label{eq:nec2}.
\end{align}
\end{proposition}
The conditions in Eq.~\ref{eq:nec1} and Eq.~\ref{eq:nec2} follow directly via Jensen's inequality applied to the LHS and RHS of Eq.~\ref{eq:nec_suf} respectively. 
Here, we note that estimates for the expectations in Eqs.~\ref{eq:nec1}-\ref{eq:nec2} based on Monte Carlo averaging of (log-) importance weights are unbiased.

\begin{table*}[t]
\centering
\caption{Goodness-of-fit evaluation on CIFAR-10 dataset for PixelCNN++ and SNGAN. Standard errors computed over 10 runs. \textbf{Higher IS is better. Lower FID and KID scores are better.} 
}\label{tab:goodnessoffit}
\small
    \begin{tabular}{@{}llccc@{}}
    \toprule
    Model                        & Evaluation    & IS ($\uparrow$) & FID ($\downarrow$) & KID ($\downarrow$) \\ \midrule
    
    - & Reference & 11.09 $\pm$ 0.1263  & 5.20 $\pm$ 0.0533 & 0.008 $\pm$ 0.0004\\ \midrule
   PixelCNN++                     &  Default (no debiasing)   &   5.16 $\pm$ 0.0117      & 58.70 $\pm$    0.0506    &    0.196  $\pm$ 0.0001    \\
          & LFIW    & \textbf{6.68} $\pm$ 0.0773& \textbf{55.83} $\pm$ 0.9695&  \textbf{0.126} $\pm$ 0.0009\\
    \midrule
          SNGAN                  &  Default (no debiasing) &   8.33$\pm $ 0.0280  &  20.40 $\pm$ 0.0747   &  0.094 $\pm$ 0.0002      \\
          & LFIW    &\textbf{ 8.57 }$\pm$ 0.0325& \textbf{17.29} $\pm$ 0.0698 &\textbf{0.073} $\pm$0.0004\\\bottomrule
    \end{tabular}
\end{table*}

\section{Application Use Cases}\label{sec:exp}
In all our experiments, the binary classifier for estimating the importance weights was a calibrated deep neural network trained to minimize the cross-entropy loss. 
The self-normalized LFIW in Eq.~\eqref{eq:self_norm_iw} worked best. 
Additional analysis on the estimators and experiment details are in Appendices~\ref{app:bv} and \ref{app:exp}.

\subsection{Goodness-of-fit testing}

In the first set of experiments, we highlight the benefits of importance weighting for a debiased evaluation of
three popularly used sample quality metrics viz. Inception Scores (IS)~\citep{salimans2016improved}, Frechet Inception Distance (FID)~\citep{heusel2017gans}, and Kernel Inception Distance (KID)~\citep{binkowski2018demystifying}. 
All these scores 
can be formally expressed as empirical expectations with respect to the model.
For all these metrics, we can simulate the population level unbiased case as a ``reference score" wherein we artificially set both the real and generated sets of samples used for evaluation as finite, disjoint sets derived from $\pdata$.

We evaluate the three metrics for two state-of-the-art models trained on the CIFAR-10 dataset viz. an autoregressive model PixelCNN++~\citep{salimans2017pixelcnn++} learned via maximum likelihood estimation and a latent variable model SNGAN~\citep{miyato2018spectral} learned via adversarial training.
For evaluating each metric, we draw 10,000 samples from the model. 
In Table~\ref{tab:goodnessoffit}, we report the metrics with and without the LFIW bias correction.
The consistent debiased evaluation of these metrics via self-normalized LFIW suggest that the SIR approximation to the importance resampled distribution (Eq.~\ref{eq:sir_approx}) is a better fit to $\pdata$.

\subsection{Data Augmentation for Multi-Class Classification}

\begin{figure*}[t]
\vspace{-0.15in}
    \centering
    \begin{subfigure}[b]{0.32\textwidth}
    \centering
    \includegraphics[width=\columnwidth]{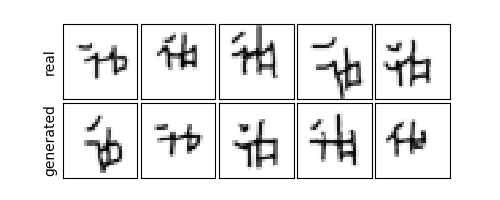}
    \caption{}
    \end{subfigure}
    \begin{subfigure}[b]{0.32\textwidth}
    \centering
    \includegraphics[width=\columnwidth]{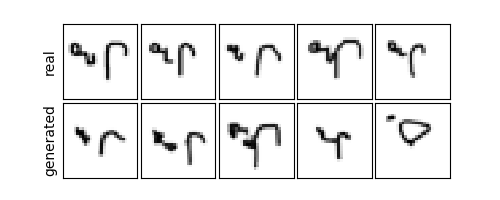}
    \caption{}
    \end{subfigure}
    \begin{subfigure}[b]{0.32\textwidth}
    \centering
    \includegraphics[width=\columnwidth]{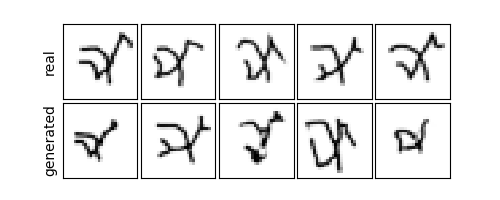}
    \caption{}
    \end{subfigure}
    \begin{subfigure}[b]{0.32\textwidth}
    \centering
    \includegraphics[width=\columnwidth]{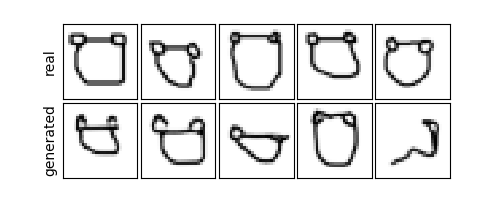}
    \caption{}
    \end{subfigure}
    \begin{subfigure}[b]{0.32\textwidth}
    \centering
    \includegraphics[width=\columnwidth]{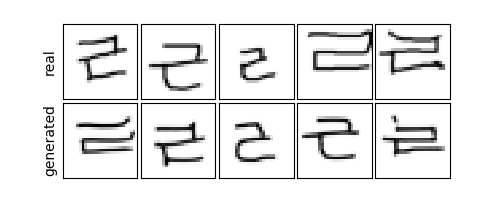}
    \caption{}
    \end{subfigure}
    \begin{subfigure}[b]{0.32\textwidth}
    \centering
    \includegraphics[width=\columnwidth]{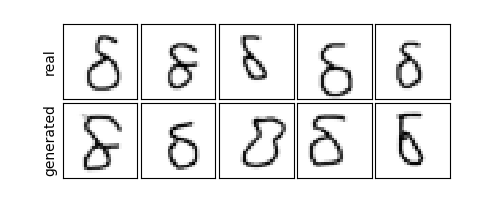}
    \caption{}
    \end{subfigure}
    \caption{Qualitative evaluation of importance weighting for data augmentation. (a-f) Top row shows held-out data samples from a specific class in Omniglot. Bottom row shows generated samples from the same class \textit{ranked in decreasing order} of importance weights.}
    \label{fig:omniglot}
    \vspace{-0.15in}
\end{figure*}

We consider data augmentation via Data Augmentation Generative Adversarial Networks (DAGAN)~\citep{antoniou2017data}.
While DAGAN was motivated by and evaluated for the task of meta-learning, it can also be applied for multi-class classification scenarios, which is the setting we consider here.
We trained a DAGAN on the Omniglot dataset of handwritten characters~\citep{lake2015human}. 
The DAGAN training procedure is described in the Appendix.
The dataset is particularly relevant because it contains 1600+ classes but only 20 examples from each class and hence, could potentially benefit from augmented data.

Once the model has been trained, it can be used for data augmentation in many ways. 
In particular, we consider ablation baselines that use various combinations of the real training data $D_{\rm{cl}}$ and generated data $D_{\rm{g}}$ for training a downstream classifier.
When the generated data $D_{\rm{g}}$ is used, we can either use the data directly with uniform weighting for all training points, or choose to importance weight (LFIW) the contributions of the individual training points to the overall loss. 
The results are shown in Table~\ref{tab:data_aug}.
While generated data ($D_{\rm{g}}$) alone cannot be used to obtain competitive performance relative to the real data ($D_{\rm{cl}}$) on this task as expected, the bias it introduces for evaluation and subsequent optimization overshadows even the naive data augmentation ($D_{\rm{cl}} + D_{\rm{g}}$).
In contrast, we can obtain significant improvements by importance weighting the generated points ($D_{\rm{cl}} + D_{\rm{g}} \textrm{ w/ LFIW}$).
\begin{table}[t]
\centering
\caption{Classification accuracy on the Omniglot dataset. Standard errors computed over 5 runs.}\label{tab:data_aug}
\vspace{0.5em}
\small
\scalebox{0.95}{
\begin{tabular}{@{}lccccc@{}}
\toprule
    Dataset &$D_{\rm{cl}}$ & $D_{\rm{g}}$ & $D_{\rm{g}} \textnormal{ w/ LFIW}$ & $D_{\rm{cl}} + D_{\rm{g}}$ & $D_{\rm{cl}} + D_{\rm{g}} \textnormal{ w/ LFIW}$  \\ \midrule
    Accuracy & 0.6603 $\pm$ 0.0012 & 0.4431 $\pm$ 0.0054 & 0.4481 $\pm$ 0.0056 & 0.6600 $\pm$ 0.0040 & \textbf{0.6818} $\pm$ 0.0022\\\bottomrule
\end{tabular}
}
\vspace{-0.1in}
\end{table}

Qualitatively, we can observe the effect of importance weighting in Figure~\ref{fig:omniglot}. Here, we show true and generated samples for $6$ randomly choosen classes (a-f) in the Omniglot dataset. 
The generated samples are ranked in decreasing order of the importance weights. 
There is no way to formally test the validity of such rankings and this criteria can also prefer points which have high density under $\pdata$ but are unlikely under $\ptheta$ since we are looking at ratios.
Visual inspection suggests that the classifier is able to appropriately downweight poorer samples, as shown in Figure~\ref{fig:omniglot} (a, b, c, d - bottom right).
There are also failure modes, such as the lowest ranked generated images in Figure~\ref{fig:omniglot} (e, f - bottom right) where the classifier weights reasonable generated samples poorly relative to others.
This could be due to particular artifacts such as a tiny disconnected blurry speck in Figure~\ref{fig:omniglot} (e - bottom right)
which could be more revealing to a classifier distinguishing real and generated data.

\subsection{Model-based Off-policy Policy Evaluation}

So far, we have seen use cases where the generative model was trained on data from the same distribution we wish to use for
Monte Carlo evaluation.
We can extend our debiasing framework to more involved settings when the generative model is a building block for specifying the full data generation process, e.g., trajectory data generated via a dynamics model along with an agent policy.

In particular, we consider the setting of off-policy policy evaluation (OPE), where the goal is to evaluate policies using experiences collected from a different policy.  Formally, let $(\gS, \gA, r, P, \eta, T)$ denote an (undiscounted) Markov decision process with state space $\gS$, action space $\gA$, reward function $r$, transition $P$, initial state distribution $\eta$ and horizon $T$. 
Assume $\pi_e: \mathcal{S} \times \mathcal{A} \to [0,1]$ is a known policy that we wish to evaluate. The probability of generating a certain trajectory $\tau = \{\vs_0, \va_0, \vs_1, \va_1, ... , \vs_T, \va_T\}$ of length $T$ with policy $\pi_e$ and transition $P$ is given as:
\begin{align}
    p^\star(\tau) &= \eta(\vs_0) \prod_{t=0}^{T-1} \pi_e(\va_t|\vs_t) P(\vs_{t+1}|\vs_t, \va_t).
    \label{eqn:pstar}
\end{align}
The return on a trajectory $R(\tau)$ is the sum of the rewards across the state, action pairs in $\tau$: $R(\tau) = \sum_{t=1}^T r(\vs_t, \va_t)$, where we assume a \emph{known reward function} $r$.

We are interested in the value of a policy defined as $v(\pi_e) = \E_{\tau \sim p^\ast(\tau)}\left[R(\tau)\right]$.
Evaluating $\pi_e$ 
requires the (unknown) transition dynamics $P$. 
The dynamics model is a conditional generative model of the next states $\vs_{t+1}$ conditioned on the previous state-action pair $(\vs_t, \va_t)$. 
If we have access to historical logged data $D_\tau$ of trajectories $\tau=\{\vs_0, \va_0, \vs_1, \va_1, \ldots,\}$ from some behavioral policy $\pi_b: \mathcal{S} \times \mathcal{A} \to [0,1]$, then we can use this off-policy data to train a dynamics model $P_\theta(\vs_{t+1} \vert \vs_{t}, \va_t)$. The policy $\pi_e$ can then be evaluated under this learned dynamics model
 as $\tilde{v}(\pi_e) = \E_{\tau \sim \tilde{p}(\tau)}[R(\tau)]$, 
where $\tilde{p}$ uses $P_\theta$ instead of the true dynamics in Eq.~\eqref{eqn:pstar}.

However, the trajectories sampled with $P_\theta$ could significantly deviate from samples from $P$ due to compounding errors~\citep{ross2010efficient}. 
In order to correct for this bias, we can use likelihood-free importance weighting on entire trajectories of data.
The binary classifier $c(\vs_t, \va_t, \vs_{t+1})$ for estimating the importance weights in this case distinguishes between triples of true and generated transitions. 
For any true triple $(\vs_t, \va_t, \vs_{t+1})$  extracted from the off-policy data, the corresponding generated triple $(\vs_t, \va_t, \hat{\vs}_{t+1})$ only differs in the final transition state, i.e., $\hat{\vs}_{t+1} \sim P_\theta(\hat{\vs}_{t+1} \vert \vs_t, \va_t)$.
Such a classifier allows us to obtain the importance weights $\hat{w}(\vs_t, \va_t, \hat{\vs}_{t+1})$ for every predicted state transition $(\vs_t, \va_t, \hat{\vs}_{t+1})$.
    The importance weights for the trajectory $\tau$ can be derived from the importance weights of these individual transitions as:
    \begin{align}
    \frac{p^\star(\tau)}{\tilde{p}(\tau)} = \frac{\prod_{t=0}^{T-1} P(\vs_{t+1}|\vs_t, \va_t)}{\prod_{t=0}^{T-1} P_\theta(\vs_{t+1}|\vs_t, \va_t)} = \prod_{t=0}^{T-1} \frac{P(\vs_{t+1}|\vs_t, \va_t)}{P_\theta(\vs_{t+1}|\vs_t, \va_t)}\approx 
    \prod_{t=0}^{T-1} \hat{w}(\vs_t, \va_t, \hat{\vs}_{t+1}). 
\end{align}
Our final LFIW estimator is given as:
\begin{align}\label{eq:iw_mbope}
    \hat{v}(\pi_e) = \E_{\tau \sim \tilde{p}(\tau)}\left[\prod_{t=0}^{T-1} \hat{w}(\vs_t, \va_t, \hat{\vs}_{t+1}) \cdot R(\tau)\right].
\end{align}

\begin{table*}
    \centering
    \caption{Off-policy policy evaluation on MuJoCo tasks. Standard error is over 10 Monte Carlo estimates where each estimate contains 100 randomly sampled trajectories. %
    }
    \vspace{0.5em}
    \small
    \begin{tabular}{lcccc}
    \toprule
    Environment        & $v(\pi_e)$ (Ground truth) & $\tilde{v}(\pi_e)$ & $\hat{v}(\pi_e)$ (w/ LFIW) & $\hat{v}_{80}(\pi_e)$ (w/ LFIW) \\\midrule
    Swimmer     & $36.7 \pm 0.1$ & $100.4 \pm 3.2$ & $\textbf{25.7} \pm 3.1$ & $\textbf{47.6} \pm 4.8$ \\
    HalfCheetah & $241.7 \pm 3.56$ & $204.0 \pm 0.8$ & $217.8 \pm 4.0$ & $\textbf{219.1} \pm 1.6$ \\ %
    HumanoidStandup & $14170 \pm 53$ & $8417 \pm 28$ & $\textbf{9372} \pm 375$ & $9221 \pm 381$ \\% & $8504 \pm 74$ & $\textbf{10223} \pm 10172$ & $9972 \pm 8163$ \\
    \bottomrule
    \end{tabular}
    \label{tab:ope}
\end{table*}

\begin{figure*}[t]
    \centering
    \begin{subfigure}{0.30\textwidth}
    \includegraphics[width=\textwidth]{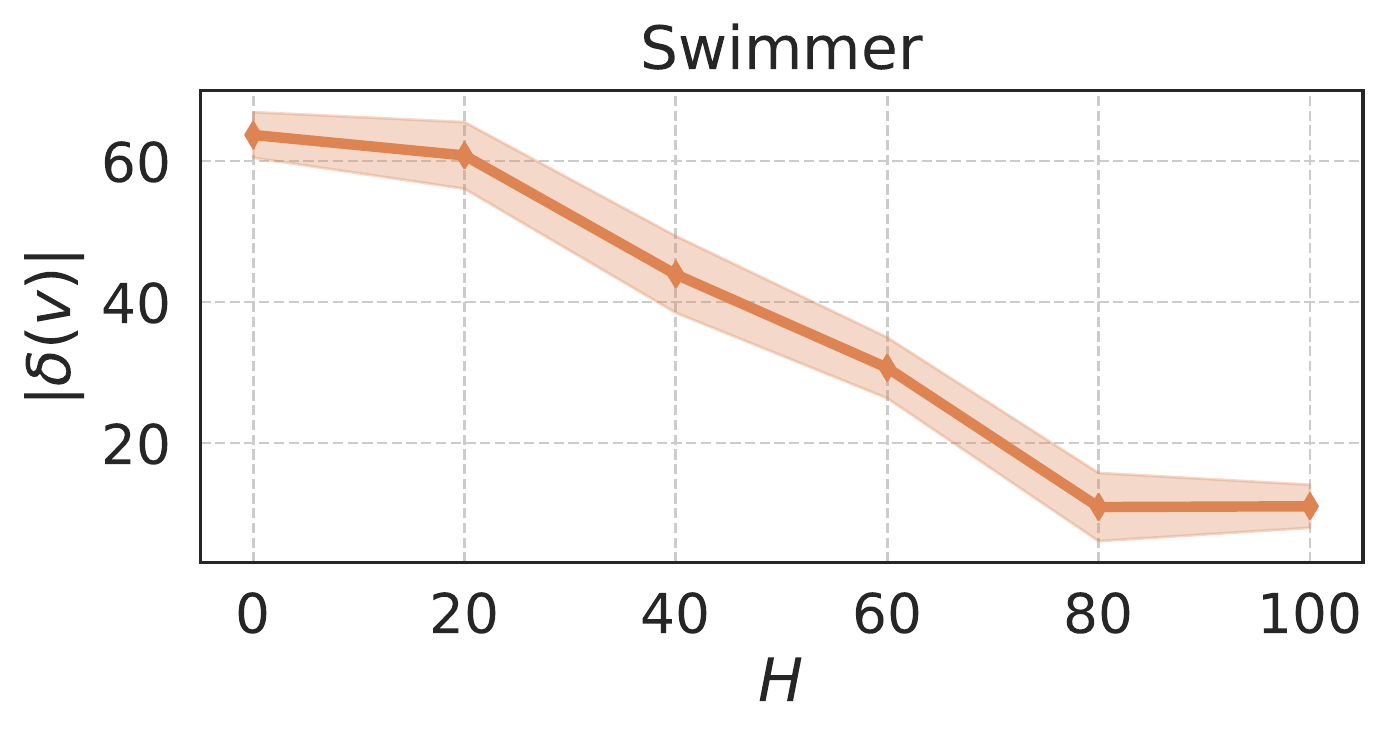}
    \end{subfigure}
    ~
    \begin{subfigure}{0.30\textwidth}
    \includegraphics[width=\textwidth]{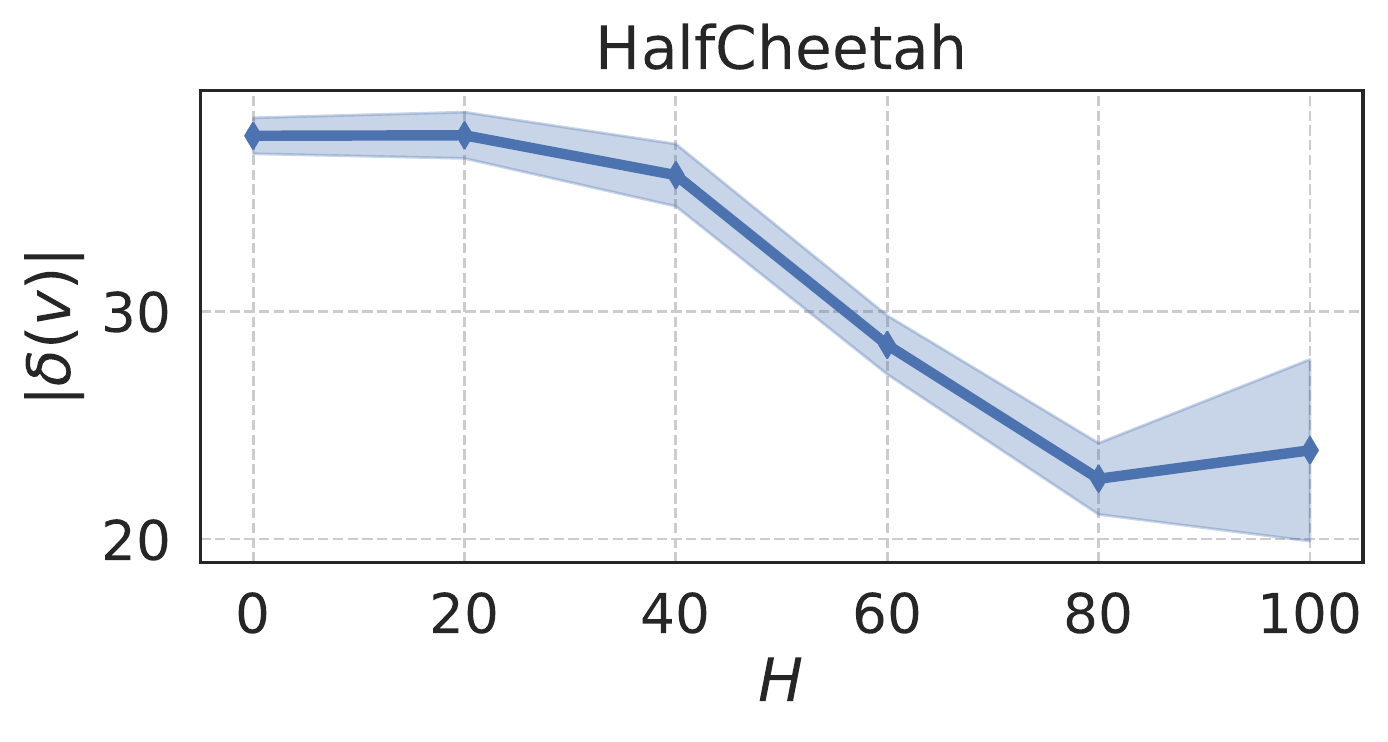}
    \end{subfigure}
    ~
    \begin{subfigure}{0.30\textwidth}
    \includegraphics[width=\textwidth]{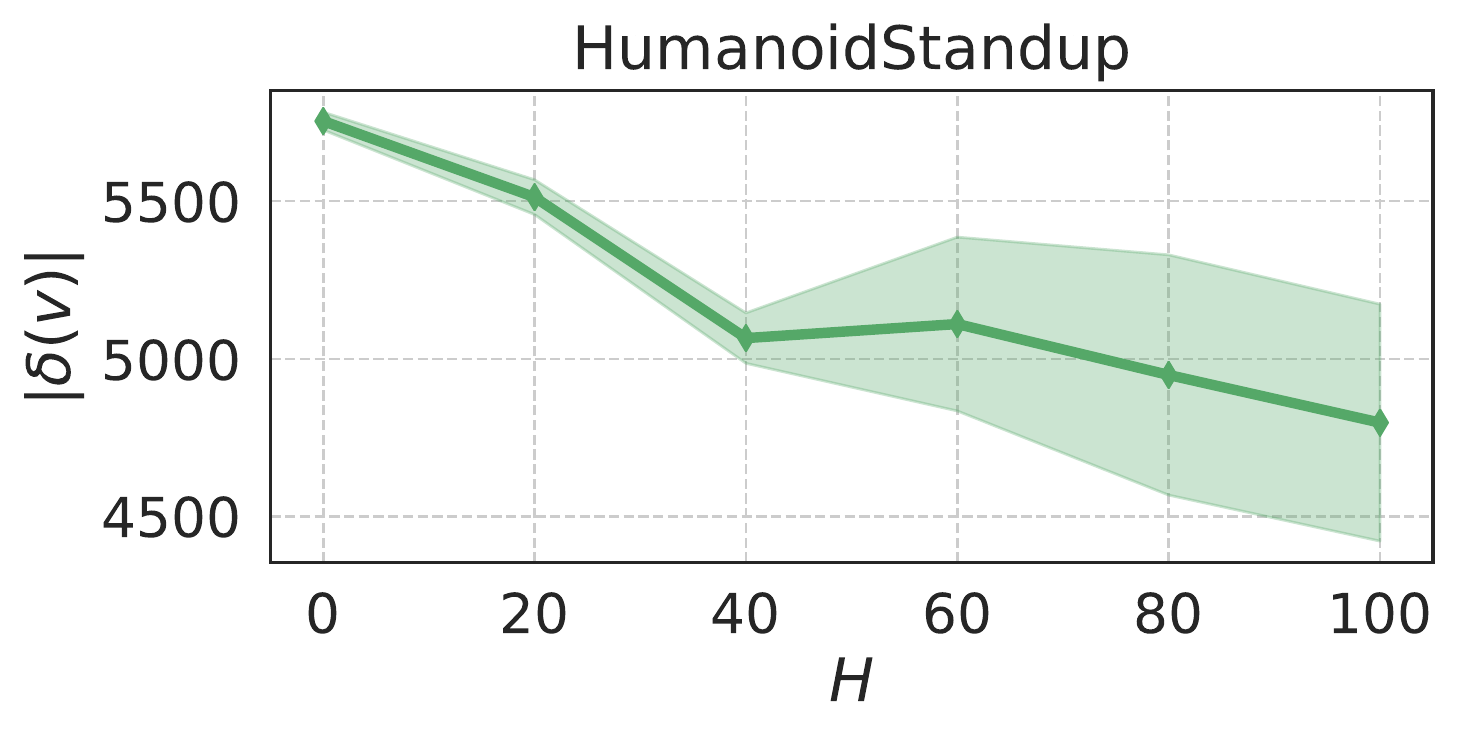}
    \end{subfigure}
    \caption{Estimation error $\delta(v) = v(\pi_e) - \hat{v}_H(\pi_e)$ for different values of $H$ (minimum 0, maximum 100). Shaded area denotes standard error over different random seeds.}
    \label{fig:ope}
\end{figure*}

We consider three continuous control tasks in the MuJoCo simulator~\citep{todorov2012mujoco} from OpenAI gym~\citep{brockman2016openai} (in increasing number of state dimensions): Swimmer, HalfCheetah and HumanoidStandup. High dimensional state spaces makes it challenging to learning a reliable dynamics model in these environments. We train behavioral and evaluation policies using Proximal Policy Optimization~\citep{schulman2017proximal} with different hyperparameters for the two policies. The dataset collected via trajectories from the behavior policy are used train a ensemble neural network dynamics model.
We the use the trained dynamics model to evaluate $\tilde{v}(\pi_e)$ and its IW version $\hat{v}(\pi_e)$, and compare them with the ground truth returns $v(\pi_e)$. Each estimation is averaged over a set of 100 trajectories with horizon $T =100$. Specifically, for $\hat{v}(\pi_e)$, we also average the estimation over 10 classifier instances trained with different random seeds on different trajectories. We further consider performing IW over only the first $H$ steps, and use uniform weights for the remainder, which we denote as $\hat{v}_H(\pi_e)$. This allow us to interpolate between $\tilde{v}(\pi_e) \equiv \hat{v}_0(\pi_e)$ and $\hat{v}(\pi_e) \equiv \hat{v}_T(\pi_e)$. Finally, as in the other experiments, we used the self-normalized 
variant (Eq.~\eqref{eq:self_norm_iw}) of the importance weighted estimator in Eq.~\eqref{eq:iw_mbope}.

 We compare the policy evaluations under different environments in Table~\ref{tab:ope}. These results show that the rewards estimated with the trained dynamics model differ from the ground truth by a large margin.
By importance weighting the trajectories, we obtain much more accurate policy evaluations. As expected, we also see that while LFIW leads to higher returns on average, the imbalance in trajectory importance weights due to the multiplicative weights of the state-action pairs can lead to higher variance in the importance weighted returns.
In Figure~\ref{fig:ope}, we demonstrate that policy evaluation becomes more accurate as more timesteps are used for LFIW evaluations, until around $80 - 100$ timesteps and thus empirically validates the benefits of importance weighting using a classifier. Given that our estimates have a large variance,
it would be worthwhile to compose our approach with other variance reduction techniques such as (weighted) doubly robust estimation in future work~\cite{farajtabar2018more}, as well as incorporate these estimates within a framework such as MAGIC to further blend with model-free OPE~\citep{thomas2016data}.
In Appendix~\ref{app:stepwise_lfiw}, we also consider a stepwise LFIW estimator for MBOPE which applies importance weighting at the level of every decision as opposed to entire trajectories.

\textbf{Overall.} Across all our experiments, we observe that importance weighting the generated samples leads to uniformly better results, whether in terms of evaluating the quality of samples, or their utility in downstream tasks. Since the technique is a black-box wrapper around any generative model, we expect this to benefit a diverse set of tasks in follow-up works. 

However, there is also some caution to be exercised with these techniques as evident from the results of Table~\ref{tab:goodnessoffit}. 
Note that in this table, the confidence intervals (computed using the reported standard errors) around the model scores after importance weighting still do not contain the reference scores obtained from the true model. 
This would not have been the case if our debiased estimator was completely unbiased and this observation reiterates our earlier claim that LFIW is reducing bias, as opposed to completely eliminating it. 
Indeed, when such a mismatch is observed, it 
is a good diagnostic to either 
learn more powerful classifiers to better approximate the Bayes optimum, or find additional data from $\pdata$ in case the generative model fails the full support assumption. 

\section{Related Work \& Discussion}\label{sec:discussion}

Density ratios enjoy widespread use across machine learning e.g., for handling covariate shifts, class imbalance etc.~\citep{sugiyama2012density,byrd2018weighted}. 
In generative modeling, estimating these ratios via binary classifiers is frequently used for defining learning objectives and two sample tests~\cite{mohamed2016learning,rosca2017variational,gretton2007kernel,bowman2015generating,lopez2016revisiting,danihelka2017comparison,rosca2017variational,im2018quantitatively,gulrajani2018towards}.
In particular, such classifiers have been used to define learning frameworks such as generative adversarial networks~\citep{goodfellow2014generative, nowozin2016f}, likelihood-free Approximate Bayesian Computation (ABC)~\citep{gutmann2012noise} and earlier work in unsupervised-as-supervised learning~\citep{friedman2001elements} and noise contrastive estimation~\citep{gutmann2012noise} among others. 
Recently, \cite{diesendruck2018importance} used importance weighting to reweigh datapoints based on differences in training and test data distributions i.e., \textit{dataset} bias.
The key difference is that these works are explicitly interested in \textit{learning} the parameters of a generative model. 
In contrast, we use the binary classifier for estimating importance weights to correct for the \textit{model} bias of any \textit{fixed} generative model.

Recent concurrent works~\cite{turner2018metropolis,azadi2018discriminator,tao2018chi} use MCMC and rejection sampling to explicitly transform or reject the generated samples.
These methods require extra computation beyond training a classifier, in rejecting the samples or running Markov chains to convergence, unlike the proposed importance weighting strategy.
For many model-based Monte Carlo evaluation usecases (e.g., data augmentation, MBOPE), this extra computation is unnecessary.
If samples or density estimates are explicitly needed from the induced resampled distribution, we presented a particle-based approximation to the induced density where the number of particles is a tunable knob allowing for trading statistical accuracy with computational efficiency.
Finally, we note resampling based techniques have been extensively studied in the context of improving variational approximations for latent variable generative models~\citep{burda2015importance, salimans2015markov, naesseth2017variational,grover2018variational}.

\section{Conclusion}

We identified bias with respect to a target data distribution as a fundamental challenge restricting the use of deep generative models as proposal distributions for Monte Carlo evaluation. 
We proposed a bias correction framework based on importance sampling. The importance weights are learned in a likelihood-free fashion via a binary classifier.
Empirically, we find the bias correction to be useful across a surprising variety of tasks including goodness-of-fit sample quality tests, data augmentation, and model-based off-policy policy evaluation.
The ability to characterize the bias of a deep generative model is an important step towards using these models to guide decisions in high-stakes applications under uncertainty~\citep{gal2016dropout,lakshminarayanan2017simple}, such as healthcare~\citep{komorowski2016markov,zhou2016detecting,raghu2017continuous} and robust anomaly detection~\citep{nalisnick2018deep,choi2018generative}.

\section*{Acknowledgments}
This project was initiated when AG was an intern at Microsoft Research. We are thankful to Daniel Levy, Rui Shu, Yang Song, and members of the Reinforcement Learning, Deep Learning, and Adaptive Systems and Interaction groups at Microsoft Research for helpful discussions and comments on early drafts.
This research was supported by NSF (\#1651565, \#1522054, \#1733686), ONR, AFOSR (FA9550-19-1-0024), and FLI.

\bibliographystyle{unsrtnat}
\bibliography{refs}
\clearpage
\appendix
\section*{Appendices}

\section{Confidence Intervals via Bootstrap}\label{app:confidence_intervals}

Bootstrap is a widely-used tool in statistics for deriving confidence intervals by fitting ensembles of models on resampled data points. 
If the dataset is finite e.g., $\train$, then the bootstrapped dataset is obtained via random sampling \textit{with replacement} and confidence intervals are estimated via the \textit{empirical bootstrap}. 
For a parametric model generating the dataset e.g., $\ptheta$, a fresh bootstrapped dataset is resampled from the model and confidence intervals are estimated via the \textit{parametric bootstrap}. See \cite{efron1994introduction} for a detailed review. 
In training a binary classifier, we can estimate the confidence intervals by retraining the classifier on a fresh sample of points from $\ptheta$ and a resampling of the training dataset $\train$ (with replacement). 
Repeating this process over multiple runs and then taking a suitable quantile gives us the corresponding confidence intervals.

\section{Bias-Variance of Different LFIW estimators}\label{app:bv}

As discussed in Section~\ref{sec:practical_lfiw}, bias reduction using LFIW can suffer from issues where the importance weights are too small due to highly confident predictions of the binary classifier.
Across a batch of Monte Carlo samples, this can increase the corresponding variance.
Inspired from the importance sampling literature, we proposed additional mechanisms to mitigate this additional variance at the cost of reduced debiasing in Eqs.~(\ref{eq:self_norm_iw}-\ref{eq:clipped_lfiw}).
We now look at the empirical bias-variance trade-off of these different estimators via a simple experiment below.

Our setup follows the goodness-of-fit testing experiments in Section~\ref{sec:exp}.
The statistics we choose to estimate is simply are the 2048 activations of the prefinal layer of the Inception Network, averaged across the test set of $10,000$ samples of CIFAR-10.

That is, the true statistics $\vs=\{s_1, s_2, \cdots, s_{2048}\}$ are given by:
\begin{align}
    s_j = \frac{1}{\vert \test \vert}\sum_{\vx \in \test} a_j(\vx)
\end{align}
where $a_j$ is the $j$-th prefinal layer activation of the Inception Network. 
Note that set of statistics $\vs$ is fixed (computed once on the test set).

To estimate these statistics, we will use different estimators.
For example, the default estimator involving no reweighting is given as:
\begin{align}\label{eq:default_lfiw_bv}
    \hat{s}_j = \frac{1}{T}\sum_{i=1}^T a_j(\vx)
\end{align}
where $\vx \sim \ptheta$.

Note that $\hat{s}_j$ is a random variable since it depends on the $T$ samples drawn from $\ptheta$.
Similar to Eq.~\eqref{eq:default_lfiw_bv}, other variants of the LFIW estimators proposed in Section~\ref{sec:practical_lfiw} can be derived using Eqs.~(\ref{eq:self_norm_iw}-\ref{eq:clipped_lfiw}).
For any LFIW estimate $\hat{s}_j$, we can use the standard decomposition of the expected mean-squared error into terms corresponding to the (squared) bias and variance as shown below.

\begin{align}
    \E[(s_j-\hat{s_j})^2] &= s_j^2 - 2 s_j\E[\hat{s_j}]+ \E[\hat{s_j}]^2\\
    &= s_j^2 - 2 s_j\E[\hat{s_j}]+ (\E[\hat{s_j}])^2 + \E[\hat{s_j}^2] - (\E[\hat{s_j}])^2\\
    &= \underbrace{(s_j - \E[\hat{s_j}])^2}_{\textrm{Bias}^2} + \underbrace{\E[\hat{s_j}^2] - (\E[\hat{s_j}])^2}_{\textrm{Variance}}.
\end{align}

In Table~\ref{tab:bias_variance_10k}, we report the bias and variance terms of the estimators averaged over 10 draws of $T=10,0000$ samples and further averaging over all $2048$ statistics corresponding to $\vs$.
We observe that self-normalization performs consistently well and is the best or second best in terms of bias and MSE in all cases.
The flattened estimator with no debiasing (corresponding to $\alpha=0$) has lower bias and higher variance than the self-normalized estimator. 
Amongst the flattening estimators, lower values of $\alpha$ seem to provide the best bias-variance trade-off.
The clipped estimators do not perform well in this setting, with lower values of $\beta$ slightly preferable over larger values.
We repeat the same experiment with $T=5,000$ samples and report the results in Table~\ref{tab:bias_variance_5k}. 
While the variance increases as expected (by almost an order of magnitude), the estimator bias remains roughly the same.

\begin{table*}[t]
\centering
\caption{Bias-variance analysis for PixelCNN++ and SNGAN when $ T=10,000$. Standard errors over the absolute values of bias and variance evaluations are computed over the 2048 activation statistics. Lower absolute values of bias, lower variance, and lower MSE is better. 
}\label{tab:bias_variance_10k}
\vspace{0.5em}
\small
    \begin{tabular}{@{}llccc@{}}
    \toprule
    Model                        & Evaluation    & $\vert \textrm{Bias} \vert$ ($\downarrow$) & Variance ($\downarrow$) & MSE ($\downarrow$) \\ \midrule
    
   PixelCNN++                     & Self-norm &\textbf{ 0.0240} $\pm$ 0.0014 & 0.0002935 $\pm$ 7.22e-06 & \textbf{0.0046} $\pm$ 0.00031 \\
& Flattening ($\alpha=0$) & 0.0330 $\pm$ 0.0023 & 9.1e-06 $\pm$ 2.6e-07 & 0.0116 $\pm$ 0.00093 \\
& Flattening ($\alpha=0.25$) & 0.1042 $\pm$ 0.0018 & \textbf{5.1e-06} $\pm$ 1.5e-07 & 0.0175 $\pm$ 0.00138 \\
& Flattening ($\alpha=0.5$) & 0.1545 $\pm$ 0.0022 & 8.4e-06 $\pm$ 3.7e-07 & 0.0335 $\pm$ 0.00246 \\
& Flattening ($\alpha=0.75$) & 0.1626 $\pm$ 0.0022 & 3.19e-05 $\pm$ 2e-06 & 0.0364 $\pm$ 0.00259 \\
& Flattening ($\alpha=1.0$) & 0.1359 $\pm$ 0.0018 & 0.0002344 $\pm$ 1.619e-05 & 0.0257 $\pm$ 0.00175 \\
& Clipping ($\beta=0.001$) & 0.1359 $\pm$ 0.0018 & 0.0002344 $\pm$ 1.619e-05 & 0.0257 $\pm$ 0.00175 \\
& Clipping ($\beta=0.01$) & 0.1357 $\pm$ 0.0018 & 0.0002343 $\pm$ 1.618e-05 & 0.0256 $\pm$ 0.00175 \\
& Clipping ($\beta=0.1$) & 0.1233 $\pm$ 0.0017 & 0.000234 $\pm$ 1.611e-05 & 0.0215 $\pm$ 0.00149 \\
& Clipping ($\beta=1.0$) & 0.1255 $\pm$ 0.0030 & 0.0002429 $\pm$ 1.606e-05 & 0.0340 $\pm$ 0.00230 \\
   \midrule
          SNGAN       

& Self-norm & 0.0178 $\pm$ 0.0008 & 1.98e-05 $\pm$ 5.9e-07 & 0.0016 $\pm$ 0.00023 \\
& Flattening ($\alpha=0$) & 0.0257 $\pm$ 0.0010 & 9.1e-06 $\pm$ 2.3e-07 & 0.0026 $\pm$ 0.00027 \\
& Flattening ($\alpha=0.25$) & \textbf{0.0096} $\pm$ 0.0007 & \textbf{8.4e-06} $\pm$ 3.1e-07 & \textbf{0.0011} $\pm$ 8e-05 \\
& Flattening ($\alpha=0.5$) & 0.0295 $\pm$ 0.0006 & 1.15e-05 $\pm$ 6.4e-07 & 0.0017 $\pm$ 0.00011 \\
& Flattening ($\alpha=0.75$) & 0.0361 $\pm$ 0.0006 & 1.93e-05 $\pm$ 1.39e-06 & 0.002 $\pm$ 0.00012 \\
& Flattening ($\alpha=1.0$) & 0.0297 $\pm$ 0.0005 & 3.76e-05 $\pm$ 3.08e-06 & 0.0015 $\pm$ 7e-05 \\
& Clipping ($\beta=0.001$) & 0.0297 $\pm$ 0.0005 & 3.76e-05 $\pm$ 3.08e-06 & 0.0015 $\pm$ 7e-05 \\
& Clipping ($\beta=0.01$) & 0.0297 $\pm$ 0.0005 & 3.76e-05 $\pm$ 3.08e-06 & 0.0015 $\pm$ 7e-05 \\
& Clipping ($\beta=0.1$) & 0.0296 $\pm$ 0.0005 & 3.76e-05 $\pm$ 3.08e-06 & 0.0015 $\pm$ 7e-05 \\
& Clipping ($\beta=1.0$) & 0.1002 $\pm$ 0.0018 & 3.03e-05 $\pm$ 2.18e-06 & 0.0170 $\pm$ 0.00171 

    \end{tabular}
\end{table*}

\begin{table*}[ht]
\centering
\caption{Bias-variance analysis for PixelCNN++ and SNGAN when $T=5,000$. Standard errors over the absolute values of bias and variance evaluations are computed over the 2048 activation statistics. Lower absolute values of bias, lower variance, and lower MSE is better. 
}\label{tab:bias_variance_5k}
\vspace{0.5em}
\small
    \begin{tabular}{@{}llccc@{}}
    \toprule
    Model                        & Evaluation    & $\vert \textrm{Bias} \vert$ ($\downarrow$) & Variance ($\downarrow$) & MSE ($\downarrow$) \\ \midrule
    
   PixelCNN++                   & Self-norm & \textbf{0.023} $\pm$ 0.0014 & 0.0005086 $\pm$ 1.317e-05 & \textbf{0.0049} $\pm$ 0.00033 \\
& Flattening ($\alpha=0$) & 0.0330 $\pm$ 0.0023 & \textbf{1.65e-05} $\pm$ 4.6e-07 & 0.0116 $\pm$ 0.00093 \\
& Flattening ($\alpha=0.25$) & 0.1038 $\pm$ 0.0018 & 9.5e-06 $\pm$ 3e-07 & 0.0174 $\pm$ 0.00137 \\
& Flattening ($\alpha=0.5$) & 0.1539 $\pm$ 0.0022 & 1.74e-05 $\pm$ 8e-07 & 0.0332 $\pm$ 0.00244 \\
& Flattening ($\alpha=0.75$) & 0.1620 $\pm$ 0.0022 & 6.24e-05 $\pm$ 3.83e-06 & 0.0362 $\pm$ 0.00256 \\
& Flattening ($\alpha=1.0$) & 0.1360 $\pm$ 0.0018 & 0.0003856 $\pm$ 2.615e-05 & 0.0258 $\pm$ 0.00174 \\
& Clipping ($\beta=0.001$) & 0.1360 $\pm$ 0.0018 & 0.0003856 $\pm$ 2.615e-05 & 0.0258 $\pm$ 0.00174 \\
& Clipping ($\beta=0.01$) & 0.1358 $\pm$ 0.0018 & 0.0003856 $\pm$ 2.615e-05 & 0.0257 $\pm$ 0.00173 \\
& Clipping ($\beta=0.1$) & 0.1234 $\pm$ 0.0017 & 0.0003851 $\pm$ 2.599e-05 & 0.0217 $\pm$ 0.00148 \\
& Clipping ($\beta=1.0$) & 0.1250 $\pm$ 0.0030 & 0.0003821 $\pm$ 2.376e-05 & 0.0341 $\pm$ 0.00232 \\
   \midrule
          SNGAN       

& Self-norm & 0.0176 $\pm$ 0.0008 & 3.88e-05 $\pm$ 9.6e-07 & 0.0016 $\pm$ 0.00022 \\
& Flattening ($\alpha=0$) & 0.0256 $\pm$ 0.0010 & 1.71e-05 $\pm$ 4.3e-07 & 0.0027 $\pm$ 0.00027 \\
& Flattening ($\alpha=0.25$) & \textbf{0.0099} $\pm$ 0.0007 & \textbf{1.44e-05} $\pm$ 3.7e-07 & \textbf{0.0011} $\pm$ 8e-05 \\
& Flattening ($\alpha=0.5$) & 0.0298 $\pm$ 0.0006 & 1.62e-05 $\pm$ 5.3e-07 & 0.0017 $\pm$ 0.00012 \\
& Flattening ($\alpha=0.75$) & 0.0366 $\pm$ 0.0006 & 2.38e-05 $\pm$ 1.11e-06 & 0.0021 $\pm$ 0.00012 \\
& Flattening ($\alpha=1.0$) & 0.0302 $\pm$ 0.0005 & 4.56e-05 $\pm$ 2.8e-06 & 0.0015 $\pm$ 7e-05 \\
& Clipping ($\beta=0.001$) & 0.0302 $\pm$ 0.0005 & 4.56e-05 $\pm$ 2.8e-06 & 0.0015 $\pm$ 7e-05 \\
& Clipping ($\beta=0.01$) & 0.0302 $\pm$ 0.0005 & 4.56e-05 $\pm$ 2.8e-06 & 0.0015 $\pm$ 7e-05 \\
& Clipping ($\beta=0.1$) & 0.0302 $\pm$ 0.0005 & 4.56e-05 $\pm$ 2.81e-06 & 0.0015 $\pm$ 7e-05 \\
& Clipping ($\beta=1.0$) & 0.1001 $\pm$ 0.0018 & 5.19e-05 $\pm$ 2.81e-06 & 0.0170 $\pm$ 0.0017 
    \end{tabular}
\end{table*}

\section{Additional Experimental Details}\label{app:exp}
\subsection{Calibration}

\begin{figure}[ht]
    \centering
    \includegraphics[height=0.35\textwidth]{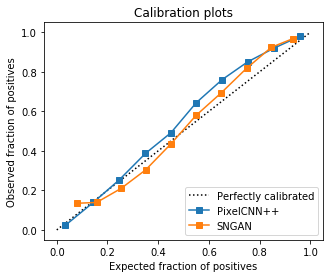}
    \caption{Calibration of classifiers for density ratio estimation.}
    \label{fig:calibration}
\end{figure}
We found in all our cases that the binary classifiers used for training the model were highly calibrated by default and did not require any further recalibration.
See for instance the calibration of the binary classifier used for goodness-of-fit experiments in Figure~\ref{fig:calibration}.
We performed the analysis on a held-out set of real and generated samples and used $10$ bins for computing calibration statistics.

We believe the default calibration behavior is largely due to the fact that our \textit{binary} classifiers distinguishing real and fake data do not require very complex neural networks architectures and training tricks that lead to miscalibration for \textit{multi-class} classification. 
As shown in \citep{niculescu2005predicting}, shallow networks are well-calibrated and \cite{guo2017calibration} further argue that a major reason for miscalibration is the use of a softmax loss typical for multi-class problems.

\subsection{Synthetic experiment}
 The classifier used in this case is a multi-layer perceptron with a single hidden layer of 100 units and has been trained to minimize the cross-entropy loss by first order optimization methods.
 The dataset used for training the classifier consists of an equal number of samples (denoted as $n$ in Figure~\ref{fig:toy}) drawn from the generative model and the data distribution.

\subsection{Goodness-of-fit testing}

We used the Tensorflow implementation of Inception Network~\citep{abadi2016tensorflow} to ensure the sample quality metrics are comparable with prior work.
For a semantic evaluation of difference in sample quality, this test is performed in the feature space of a pretrained classifier, such as the prefinal activations of the Inception Net~\citep{szegedy2016rethinking}. 
For example, the Inception score for a generative model $\ptheta$ given a classifier $d(\cdot)$ can be expressed as:
\begin{align*}
     \textnormal{IS} &= \exp(\E_{\vx \sim \ptheta}[\textnormal{KL}(d(y \vert \vx), d(y))]).
\end{align*}
The FID score is another metric which unlike the Inception score also takes into account real data from $\pdata$. Mathematically, the FID between sets $S$ and $R$ sampled from distributions $\ptheta$ and $\pdata$ respectively, is defined as:
\begin{align*}
\fid(S, R) = \Vert \mu_S - \mu_R \Vert_2^2 + \textnormal{Tr}(\Sigma_S + \Sigma_R - 2\sqrt{\Sigma_S \Sigma_R})
\end{align*}
where $(\mu_S, \Sigma_S)$ and $(\mu_R, \Sigma_R)$ are the empirical means and covariances computed based on $S$ and $R$ respectively. 
Here, $S$ and $R$ are sets of datapoints from $\ptheta$ and $\pdata$. 
In a similar vein, KID compares statistics between samples in a feature space defined via a combination of kernels and a pretrained classifier.
The standard kernel used is a radial-basis function kernel with a fixed bandwidth of $1$.
As desired, the score is optimized when the data and model distributions match.

We used the open-sourced model implementations of PixelCNN++~\citep{salimans2016improved} and SNGAN~\citep{miyato2018spectral}.
Following the observation by \cite{lopez2016revisiting}, we found that training a binary classifier on top of the feature space of any pretrained image classifier was useful for removing the low-level artifacts in the generated images in classifying an image as real or fake.
We hence learned a multi-layer perceptron (with a single hidden layer of $1000$ units) on top of the $2048$ dimensional feature space of the Inception Network. 
Learning was done using the Adam optimizer with the default hyperparameters with a learning rate of $0.001$ and a batch size of $64$.
We observed relatively fast convergence for training the binary classifier (in less than $20$ epochs) on both PixelCNN++ and SNGAN generated data and the best validation set accuracy across the first $20$ epochs was used for final model selection.

\subsection{Data Augmentation}

Our codebase was implemented using the PyTorch library~\citep{paszke2017automatic}.
We built on top of the open-source implementation of DAGAN\footnote{\url{https://github.com/AntreasAntoniou/DAGAN.git}}~\citep{antoniou2017data}.

A DAGAN learns to augment data by training a conditional generative model $G_\theta: \calX \times \calZ \to \calX$ 
based on a training dataset $D_{\rm{cl}}$. 
This dataset is same as the one we used for training the generative model and the binary classifier for density ratio estimation.
The generative model is learned via a minimax game with a critic.
For any conditioning datapoint $\vx_i \in \train$ and noise vector $\vz \sim p(\vz)$, the critic learns to distinguish the generated data $G_\theta(\vx_i, \vz)$ paired along with $\vx_i$ against another pair $(\vx_i, \vx_j)$. 
Here, the point $\vx_j$ is chosen such that the points $\vx_i$ and $\vx_j$ have the same label in  $D_{\rm{cl}}$, i.e., $y_i=y_j$. 
Hence, the critic learns to classify pairs of (real, real) and (real, generated) points while encouraging the generated points to be of the same class as the point being conditioned on. 
For the generated data, the label $y$ is assumed to be the same as the class of the point that was used for generating the data. 
We refer the reader to \cite{antoniou2017data} for further details.

Given a DAGAN model, we additionally require training a binary classifier for estimating importance weights and a multi-class classifier for subsequent classification.
The architecture for both these use cases follows prior work in meta learning on Omniglot~\citep{vinyals2016matching}.
We train the DAGAN on the 1200 classes reserved for training in prior works. 
For each class, we consider a 15/5/5 split of the 20 examples for training, validation, and testing. 
Except for the final output layer, the architecture consists of 4 blocks of 3x3 convolutions and 64 filters, followed by batch normalization~\citep{szegedy2016rethinking}, a ReLU non-linearity and 2x2 max pooling.
Learning was done for 100 epochs using the Adam optimizer with default parameters and a learning rate of 0.001 with a batch size of 32.

\subsection{Model-based Off-policy Policy Evaluation}

\begin{figure}[ht]
    \centering
    \begin{subfigure}{0.2\textwidth}
    \centering
    \includegraphics[height=\textwidth]{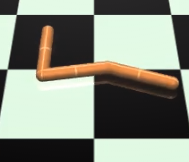}
    \caption{Swimmer}
    \end{subfigure}
    ~\hspace{0.03\textwidth}
    \begin{subfigure}{0.2\textwidth}
    \centering
    \includegraphics[height=\textwidth]{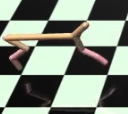}
    \caption{HalfCheetah}
    \end{subfigure}
    ~\hspace{0.03\textwidth}
    \begin{subfigure}{0.2\textwidth}
    \centering
    \includegraphics[height=\textwidth]{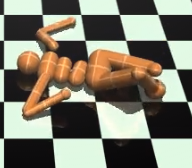}
    \caption{HumanoidStandup}
    \end{subfigure}
    \caption{Environments in OPE experiments.}
    \label{fig:ope-envs}
\end{figure}

 For this set of experiments, we used Tensorflow~\citep{abadi2016tensorflow} and OpenAI baselines\footnote{\url{https://github.com/openai/baselines.git}}~\citep{dhariwal2017openai}.
We evaluate over three envionments viz. Swimmer, HalfCheetah, and HumanoidStandup (Figure~\ref{fig:ope-envs}. 
Both HalfCheetah and Swimmer rewards the agent for gaining higher horizontal velocity; HumanoidStandup rewards the agent for gaining more height via standing up. 
In all three environments, the initial state distributions are obtained via adding small random perturbation around a certain state. 
The dimensions for state and action spaces are shown in Table~\ref{tab:ope-envs}.

\begin{table}[ht]
    \centering
    \caption{Statistics for the environments.}
    \vspace{0.5em}
    \begin{tabular}{c|cc}
        \toprule
        Environment & State dimensionality & \# Action dimensionality \\
         \midrule
        Swimmer & 8 & 2 \\
        HalfCheetah & 17 & 6 \\
        HumanoidStandup & 376 & 17 \\\bottomrule
    \end{tabular}
    \label{tab:ope-envs}
\end{table}

Our policy network has two fully connected layers with 64 neurons and tanh activations for each layer, where as our transition model / classifier has three hidden layers of 500 neurons with swish activations~\citep{ramachandran2018searching}.
We obtain our evaluation policy by training with PPO for 1M timesteps, and our behavior policy by training with PPO for 500k timesteps. Then we train the dynamics model $P_\theta$ for 100k iterations with a batch size of 128. Our classifier is trained for 10k iterations with a batch size of 250, where we concatenate $(\vs_t, \va_t, \vs_{t+1})$ into a single vector.

\begin{table*}
    \centering
    \caption{Off-policy policy evaluation on MuJoCo tasks. Standard error is over 10 Monte Carlo estimates where each estimate contains 100 randomly sampled trajectories. Here, we perform stepwise LFIW over transition triplets.
    }
    \vspace{0.5em}
    \small
    \begin{tabular}{lcccc}
    \toprule
    Environment        & $v(\pi_e)$ (Ground truth) & $\tilde{v}(\pi_e)$ & $\hat{v}(\pi_e)$ (w/ LFIW) & $\hat{v}_{80}(\pi_e)$ (w/ LFIW) \\\midrule
    Swimmer     & $36.7 \pm 0.1$ & $100.4 \pm 3.2$ & $19.4 \pm 4.3$ & $\textbf{48.3} \pm 4.0$ \\
    HalfCheetah & $241.7 \pm 3.6$ & $204.0 \pm 0.8$ & $\textbf{229.1} \pm 4.9$ & $214.9 \pm 3.9$\\
    HumanoidStandup & $14170 \pm 5.3$ & $8417 \pm 28$ & $\textbf{10612} \pm 794$ & $9950 \pm 640$ \\
    \bottomrule
    \end{tabular}
    \label{tab:ope_t}
\end{table*}

\begin{figure*}[ht]
    \centering
    \begin{subfigure}{0.30\textwidth}
    \includegraphics[width=\textwidth]{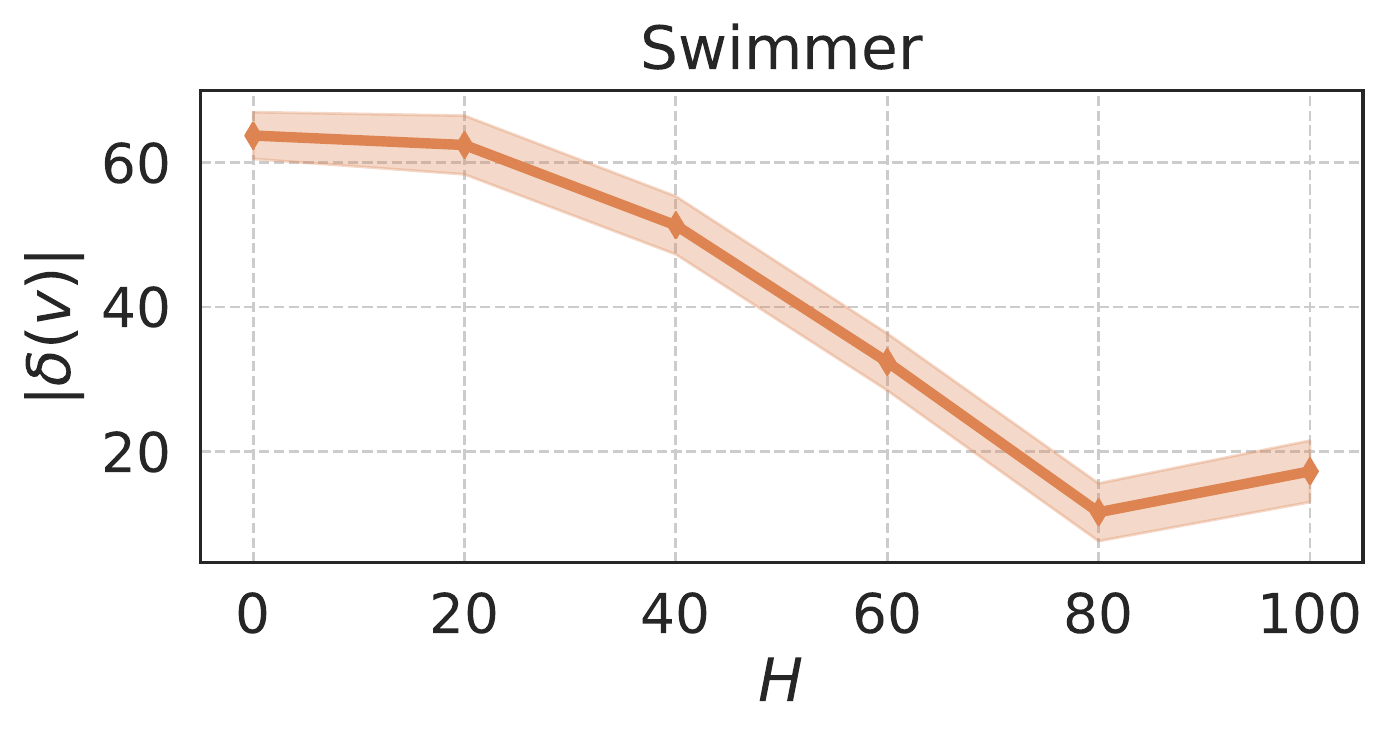}
    \end{subfigure}
    ~
    \begin{subfigure}{0.30\textwidth}
    \includegraphics[width=\textwidth]{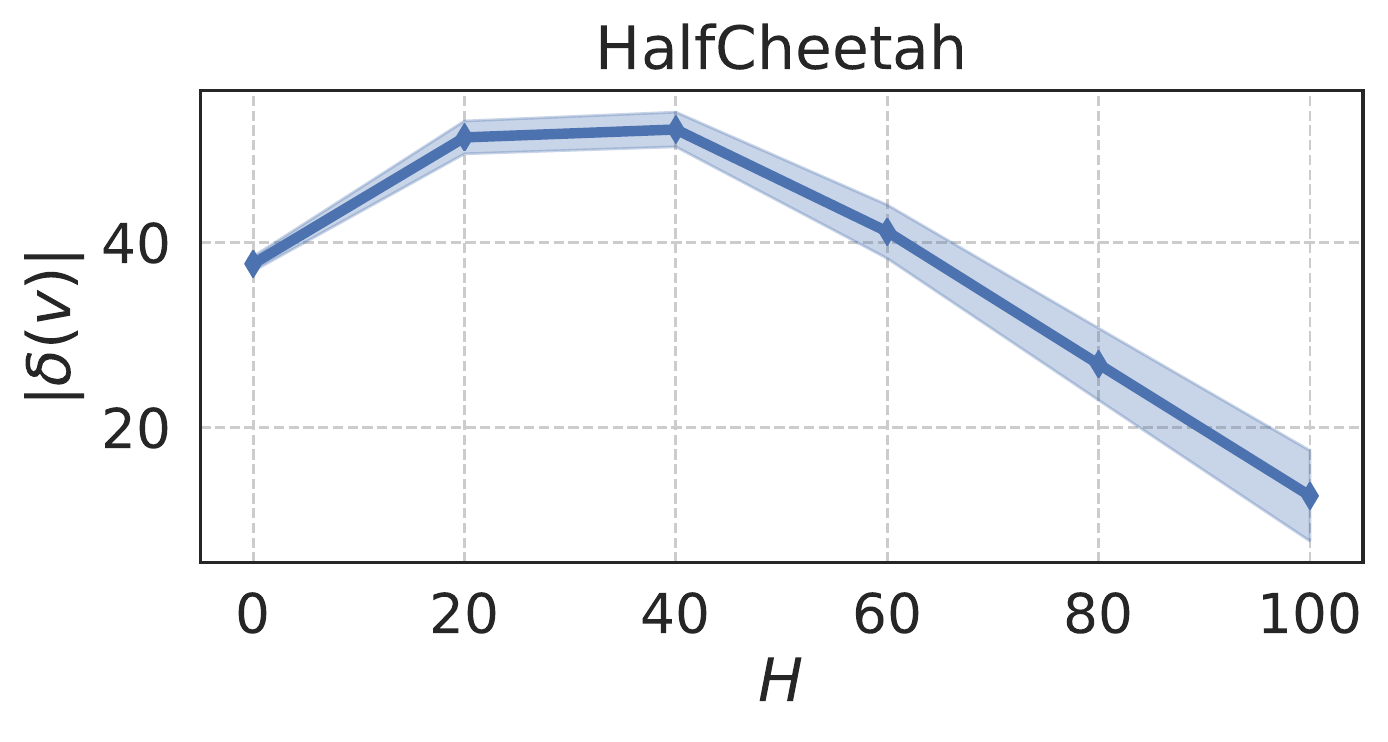}
    \end{subfigure}
    ~
    \begin{subfigure}{0.30\textwidth}
    \includegraphics[width=\textwidth]{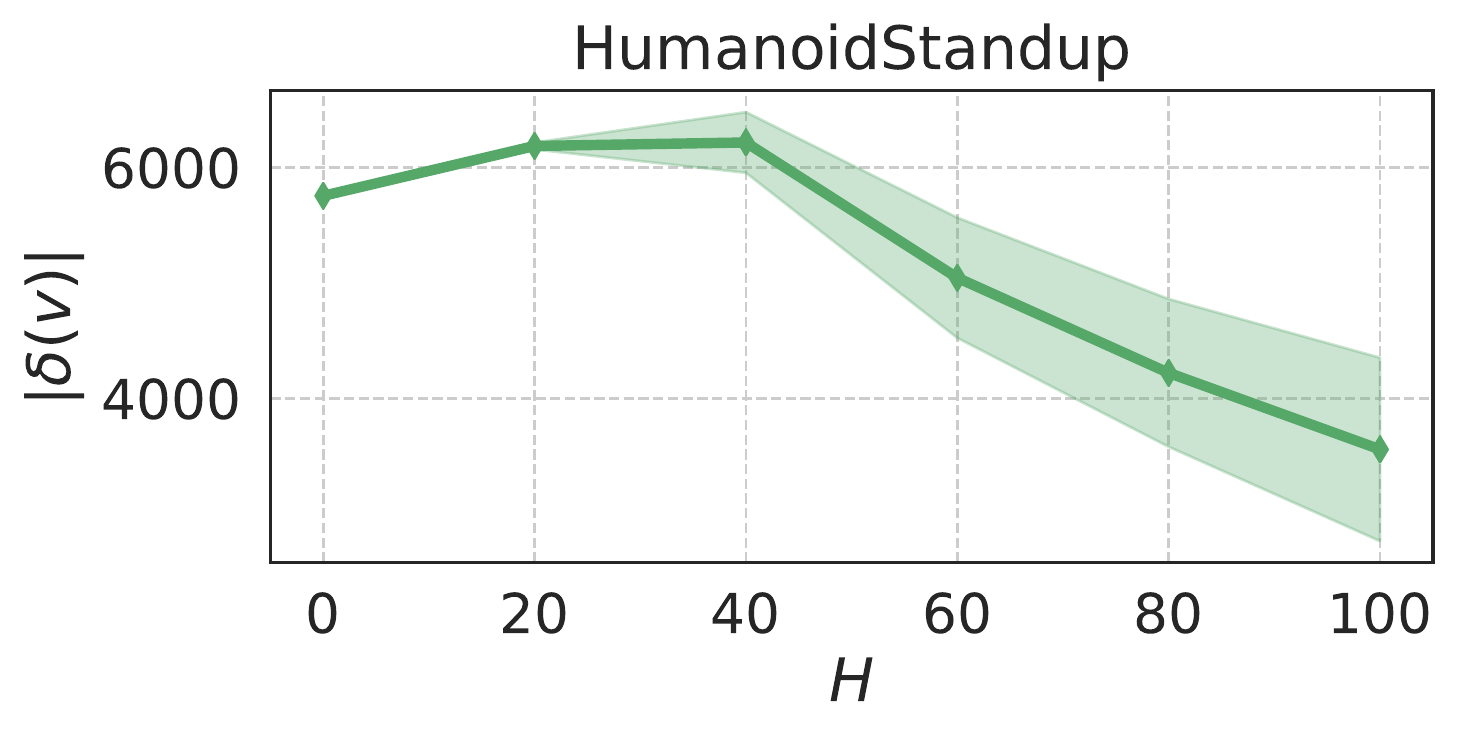}
    \end{subfigure}
    \caption{Estimation error $\delta(v) = v(\pi_e) - \hat{v}_H(\pi_e)$ for different values of $H$ (minimum 0, maximum 100). Shaded area denotes standard error over different random seeds; each seed uses 100 sampled trajectories. Here, we use LFIW over transition triplets.}
    \label{fig:ope_t}
\end{figure*}
\subsubsection{Stepwise LFIW}\label{app:stepwise_lfiw}
Here, we consider performing LFIW over the transition triplets, where each transition triplet $(\vs_{t}, \va_{t}, \vs_{t+1})$ is assigned its own importance weight.
This is in contrast to assigning a single importance weight for the entire trajectory, obtained by multiplying the importance weights of all transitions in the trajectory.
The importance weight for a transition triplet is defined as:
\begin{align}
    \frac{p^\star(\vs_{t}, \va_{t}, \vs_{t+1})}{\tilde{p}(\vs_{t}, \va_{t}, \vs_{t+1})} \approx \hat{w}(\vs_{t}, \va_{t}, \vs_{t+1}),
\end{align}
so the corresponding LFIW estimator is given as
\begin{align}
    \hat{v}(\pi_e) = \E_{\tau \sim \tilde{p}(\tau)}\left[\sum_{t=0}^{T-1} \hat{w}(\vs_t, \va_t, \hat{\vs}_{t+1}) \cdot r(\vs_t, \va_t)\right].
\end{align}
We describe this as the ``stepwise" LFIW approach for off-policy policy evaluation. We perform self-normalization over the weights of each triplet.

From the results in Table~\ref{tab:ope_t} and Figure~\ref{fig:ope_t}, stepwise LFIW also reduces bias for OPE compared to without LFIW. Compared to the ``trajectory based" LFIW described in Eq.~(\ref{eq:iw_mbope}), the stepwise estimator has slightly higher variance and weaker performance for $H = 20, 40$, but outperforms the trajectory level estimators when $H = 100$ on HalfCheetah and HumanoidStandup environments.
\end{document}